\documentclass[twoside,leqno,twocolumn]{article}
\usepackage{ltexpprt}
\usepackage{caption,booktabs, dblfloatfix}
\usepackage{color}
\usepackage{makeidx}  % allows for indexgeneration
\usepackage{multirow}      
\usepackage{subfig,graphicx}
\usepackage{epstopdf}
\usepackage{subfig}
\usepackage{amsmath,dsfont}
\usepackage{amsfonts}
\usepackage{graphicx,epstopdf}
\usepackage{enumitem}
\usepackage{color}
\usepackage{bbding}
\usepackage{wasysym}
\usepackage{colonequals}
\usepackage{psfrag}
\usepackage{mathrsfs}
\usepackage{lscape}
\usepackage{cite}
\usepackage{setspace}
\usepackage{pifont,textcomp,bbm}
\usepackage{algorithm, algpseudocode}
\usepackage{dsfont}
\usepackage{amsmath,calc}
\usepackage{graphicx}
\newtheorem{definition}{Definition}
\newtheorem{lm}{Lemma}
\begin{document}
\title{Multi-Value Rule Sets}
\author{Tong Wang \\ University of Iowa \\tong-wang@uiowa.edu}
\date{}

\maketitle

% Copyright Statement
% When submitting your final paper to a SIAM proceedings, it is requested that you include 
% the appropriate copyright in the footer of the paper.  The copyright added should be 
% consistent with the copyright selected on the copyright form submitted with the paper.
% Please note that "20XX" should be changed to the year of the meeting.

% Default Copyright Statement
%\fancyfoot[R]{\footnotesize{\textbf{Copyright \textcopyright\ 2018 by SIAM\\
%Unauthorized reproduction of this article is prohibited}}}

\begin{abstract} \small\baselineskip=9pt
We present the Multi-vAlue Rule Set (MARS) model for interpretable classification with feature efficient presentations. MARS introduces a more generalized form of association rules that allows multiple values in a condition. Rules of this form are more concise than traditional single-valued rules in capturing and describing patterns in data. MARS mitigates the problem of dealing with continuous features and high-cardinality categorical features faced by rule-based models. Our formulation also pursues a higher efficiency of feature utilization, which reduces the cognitive load to understand the decision process. We propose an efficient inference method for learning a maximum \emph{a posteriori} model, incorporating theoretically grounded bounds to iteratively reduce the search space to improve search efficiency.  
Experiments with synthetic and real-world data demonstrate that MARS models have significantly smaller complexity and fewer features, providing better interpretability while being competitive in predictive accuracy. We conducted a usability study with human subjects and results show that MARS is the easiest to use compared with other competing rule-based models, in terms of the correct rate and response time. Overall, MARS introduces a new approach to rule-based models that balance accuracy and interpretability with feature-efficient representations.
\end{abstract}

%\vspace{-4mm}
\section{Introduction}\label{sec:intro}
% contribution1: categorical features with high cardinality
In many real-world applications of machine learning, domain experts desire interpretability of a model as much as the predictive accuracy. In some occasions, interpretability even outweighs accuracy. As opposed to ``black box'' models, interpretable models provide a way for humans to understand the decision process, which is imperative in domains such as healthcare and judiciary. Among different forms of interpretable models, we are particularly interested in rule-based models in this paper.  This type of models generate decisions based on a set of rules following simple ``if-then'' logic: if a rule (or a set of rules) is satisfied, the model outputs the corresponding decision. The set of rules can be either ordered \cite{WangRu15,li2001cmar,yin2003cpar, cohen1995fast,ynormalize_addang2016scalable} or unordered \cite{lakkaraju2016interpretable, wangbayesian,malioutov2013exact, rijnbeek2010finding}, depending on the specific model structure.

Prior rule-based models in literature are built from rules with single values in each condition \cite{WangRu15,li2001cmar,yin2003cpar, cohen1995fast,lakkaraju2016interpretable, wangbayesian,wang2017bayesian, malioutov2013exact}.  For example, $[\text{state }=\text{California}]$ \emph{AND} $[\text{marital status }=\text{married}]$, where a condition (e.g., $[\text{state }=\text{california}]$) is a pair of a feature (e.g., state) and a single value (e.g., California). However, this form of rules induce implicit limitations on the performance of a model and there are imperative problems that need to be addressed.

First, models built from single-value rules are limited in handling categorical features with a medium to high cardinality of the value set. Yet this type of features is quite common in real-world data mining applications. They often contain potentially relevant information that is too costly to present in single-value rule forms. Specifically, to capture a group of instances where multiple values are interchangeable in a rule,  multiple rules need to co-exist in a model to characterize all cases. For example, to capture a set of married or divorced people who live in California, Texas, Arizona, and Oregon,  a model needs to include eight rules, each rule being a combination of state and marital status, yielding an overly complicated and repetitive model. %This issue is not only specific to categorical features. %If numerical attributes are discretized into too many intervals, same problems occur. 
This problem is less severe fore decision trees which allow a split of the values into value sets. But decision trees are greedy which often generate models suboptimal in both predictive accuracy and interpretability.

Second, rule-based models are notoriously bad at handling continuous attributes. Most rule-based models deal with them by discretizing continuous features into pre-defined intervals and then treat them as categorical features. The intervals are usually determined subject to domain knowledge or rules of thumb, such as equal-width or equal-frequency discretization. However, improper intervals will affect the predictive performance. If the interval size is too small, the  rule will have too low a support, causing overfitting. If the interval size is too large, the  rule may become imprecise, yielding a decreased overall accuracy.  %Some methods were proposed to learn a globally optimum cut-off points for discretizing continuous features, but the problem is NP-hard, and the complexity quickly explodes even with medium sized data set \cite{goh2014box, wang2015learning}.

Finally, another important aspect that has been overlooked by previous rule-based models is the need to control the total number of unique features. The number of different entities (features) humans need to comprehend is directly associated with how easy it is to understand the model \cite{miller1956magical} since humans can only process 7$\pm$2 entities at a time. %With fewer features involved, it also becomes easier for domain experts to gain clear insights into data. 
Besides, models using fewer features are easier to implement and explain. We call them \emph{feature-efficient}.
 To give an example, we examine the following rules,  the first rule is $[\text{state }=\text{California or Texas}]$ \emph{AND} $[\text{marital status }=\text{married}]$ and the second rule is $[\text{state }=\text{California}]$ \emph{AND} $[\text{marital status }=\text{married}]$ \emph{AND} $[\text{age }\geq 45]$. Although both rules have 3 values, the former is more concise  since California is grouped with Texas so that the rule appears in 2 conditions.

Here, we propose a novel rule-based classifier, Multi-value Rule Sets (MARS), which consists of rules that allow multiple values in a condition and uses a small set of features in all rules. See two examples below. Example 1 is equivalent to 8 single-value rules. For continuous features, we treat an interval as a category and multiple values can be merged. For example, if age is discretized into 10-year intervals, and $[0,10),[10,20),[20,30),[40,50)$ are selected into the rule, they can be merged as it is in Example 2. Example 2 is also equivalent to 8 single-value rules if discretized as such.

\vspace{2mm}
\noindent\text{Example 1: }[\text{state }=\text{California }\textbf{or}\text{ Texas }\textbf{or}\text{ Arizona }\textbf{or} \text{Oregon}] \emph{ AND } [\text{marital status }=\text{married }\textbf{or}\text{ divorced}]    

\vspace{2mm}
\noindent\text{Example 2: }[$\text{age}\in [0,30)\cup [40,50)$]     
\vspace{2mm}

This form of rules has great sadvantage over single-value rules for (i) a more concise presentation of information (ii) mitigating the performance loss caused by pre-discretizing continuous features, and (iii) using a smaller number of features in the model.

We develop a Bayesian framework for learning MARS and propose a principled objective combining interpretability and predictive accuracy. The interpretability is associated with the appearance of the rule set, i.e., the set of features, and the complexity of the model, described by the number of rules and the lengths of rules. We introduce Poisson distributions to govern the complexity and apply  Dirichlet distributions to feature assignment.  Feature assignment hence takes advantage of the ``clustering'' property of Dirichlet distributions and encourages re-using existing features in a rule. Therefore, this prior model for forming a MARS promotes a \emph{small} set of \emph{short} rules using a \emph{few} features. Given a MARS, we then model the conditional likelihood as a set of Bernoulli distributions with parameters determined by the coverage of MARS. We propose an efficient inference algorithm for learning a maximum \emph{a posteriori} model where we exploit the model formulation to derive theoretical bounds for reducing computation.  
We show with experimentation that MARS  achieves prediction accuracy comparable to or better than prior art for standard datasets with significantly lower complexity and fewer  features.

\section{Related Work}
There has been a series of research on developing rule-based models for classification \cite{wangbayesian, lakkaraju2016interpretable,ishibuchi2001effect, chi1996fuzzy, pmlr-v56-Tran16, malioutov2013exact,cohen1995fast, rijnbeek2010finding}. Various structures and formats of models were proposed, from the earlier work on Classification based on Association Rules (CBA) \cite{ma1998integrating} and Repeated Incremental Pruning to Produce Error Reduction (Ripper) \cite{cohen1995fast} to more recent work on rule sets \cite{wangbayesian,lakkaraju2016interpretable, malioutov2013exact} and rule lists \cite{LethamRuMcMa15, WangRu15}. A major development along this line of work is that interpretability has been recognized and emphasized. Therefore  controlling model complexity for easier interpretation is becoming an important component in the modeling. However, all of  models above use single-value rules and are limited in expressive power, leaving redundancy in the model. In addition, learning in previous methods is mostly a two-step procedure\cite{wangbayesian, ma1998integrating, LethamRuMcMa15}, that first uses off-the-shelf data mining algorithms to generate a set of rules and then chooses a set from them to form the final model. This in practice will encounter the bottleneck of mining long rules. (Million of rules can be generated from a medium size dataset if the maximum length is set to only 3. \cite{wangbayesian}). Furthermore, few of previous work consider limiting the number of features, and it is often independent of rule mining. There has not been any work that combines rule learning and feature assignment into a unified framework.

Our work is also related to working with high-cardinality categorical data, which is very challenging to handle for data mining algorithms.  Micci-Barreca
\cite{micci2001preprocessing} presented a data-preprocessing scheme that transforms high-cardinality categorical attributes into quasi-continuous scalar attributes suited for use in regression-type models. Recent works focus more on combining multiple values into a single feature \cite{ritchie2015methods,papaxanthos2016finding}. This type of approach can only be used as a preliminary step here. Previous rule-based models did not incorporate values grouping into the model learning.

Here, we propose a unified framework that combines rule mining, feature assignment, and rule selection into a single model, optimizing a joint objective of predictive accuracy and interpretability.

\section{Multi-value Rule Sets}\label{sec:prem}
We work with standard classification data set $S$ that consists of $N$ observations $\{\mathbf{x}_n, y_n\}_{n=1}^N$. Let $\mathbf{y}$ represent the set of labels and $\mathbf{x}$ represent the set of covariates $\mathbf{x}_n$. Each observation has $J$ features, and we denote the $j^\text{th}$ feature of the $n^\text{th}$ observation as $x_{nj}$.  Let $\mathcal{V}_j$ represent a set of values the $j^\text{th}$ feature takes. This notation can simply adapt to continuous attributes by discretizing the values into intervals.

\subsection{Multi-value Rules}\label{sec:mvr}
 Now we introduce the basic components in Multi-value Rule Set model.
\begin{definition}
An item is a pair of a feature $j$ and a value $v$, where $j \in \{1,2,\cdots,J\}$ and $v \in \mathcal{V}_j$.
\end{definition}
%Given a dataset $S$, we define an item space as a set of all possible pairs of feature and its values in $S$, denoted as  $\mathcal{I}_S = \{(j, v)|j \in \{1,\cdots,J\}, v \in \mathcal{V}_j\}$.
\begin{definition}
A condition is a collection of items with the same feature $j$,  denoted as $c = (j, V)$, where $j\in \{1,2,\cdots,J\}$ and $V\subset \mathcal{V}_j$. $V$ is a set of values in the item.
\end{definition}
\begin{definition}
A multi-vale rule is a conjunction of conditions, denoted as $r = \{c_k\}_{k}$.
\end{definition}
Following the definitions, an item is the atom in a multi-value rule. It is also a special case of a condition with a single value, for example, $[\text{state }=\text{California}]$. Interchangeable values are grouped  into a value set.  In this way, a multi-value rule can easily describe the example mentioned in Section~\ref{sec:intro}. We only need one rule instead of eight, yielding a more concise presentation while preserving the information. Similarly for continuous features, multiple smaller intervals can be selected and merged into a more compact presentation. (See Example 1\&2 in Introduction.)

%[\text{state }=\text{California or Texas or Arizona or Oregon}] \emph{ AND } [\text{marital status }=\text{married or divorced}].

%    When $|V| = 1$ in all conditions, a multi-value rule is a classic single-value association rule. %The number of conditions in a rule is called the \emph{length} of the rule, denoted as $|r|$.

Now we define a classifier built from multi-value rules. By an abuse of notation, we use $r(\cdot)$ to represent a Boolean function that indicates if an observation satisfies rule $r$:
%\begin{equation}\label{eqn:rule_boolen}
$r(\cdot):\mathcal{X} \rightarrow \{0,1\}.$
%\end{equation}
Let $R$ denote a Multi-value Rule Set. 
We define a classifier $R(\cdot)$:
\begin{equation}\label{eqn:classifier}
R(\mathbf{x}) = \begin{cases} 1 & \exists r \in R, r(\mathbf{x})=1 \\  0 & \text{otherwise.} \end{cases}
\end{equation}
$\mathbf{x}$ is classified as positive if it obeys \emph{at least} one rule in $R$ and we say $\mathbf{x}$ is \emph{covered} by $r$.

%Now we present the Bayesian model for learning a MARS model from data. 

\subsection{MARS Formulation}\label{sec:model}
Our proposed framework considers two aspects of a model: 1) interpretability, characterized by a prior model for MARS, which considers the complexity (number of rules and lengths of rules) and feature assignment. 2) predictive accuracy, represented by the conditional likelihood of data given a MARS model. Both components have tunable parameters to trade off between interpretability and predictive accuracy. Now we formulate the model.

  \paragraph{Prior for MARS}
%  Interpretability of a rule set refers to the cognitive load to understand the model, which is associated with 
The prior model determines the number of rules $M$, lengths of rules $\{L_m\}_m$ and feature assignment $\{z_{m}\}_{m}$, where $m$ is the rule index and $z_m$ is a vector. We propose a  two-step process for constructing a rule set, where the first step determines the size and shape of a MARS model and the second step fills in the empty ``boxes'' with items. 

\emph{Creating empty ``boxes'' - complexity assignment} First, we draw the number of rules $M$ from a Poisson distribution. The arrival rate $\lambda_M$ is determined via a Gamma distribution with parameters $\alpha_M, \beta_M$. Second, we determine the number of items in each rule, denoted as $L_m$. $L_m \sim \text{Poisson}(\lambda_L)$, which is a Poisson distribution truncated to only allow positive outcomes. The arrival rate for this Poisson distribution, $\lambda_L$, is governed by a Gamma distribution with parameters $\alpha_L, \beta_L$.  Since we favor simpler models for interpretability purposes, we set $\alpha_L< \beta_L$ and $\alpha_M<\beta_M$ to encourage a small set of short rules. These two steps together determine the size and shape of a MARS model. We call parameters $\alpha_M,\beta_M,\alpha_L,\beta_L$ shape parameters.  This step creates empty  ``boxes'' to be filled with items in the following step and assigns overall complexity to the model.

\emph{Filling ``boxes'' - feature assignment:} Rule $m$ is a collection of $L_m$ ``boxes'', each containing an item. Let $z_{mk}$ represent the feature assigned to the $k^\text{th}$ box in rule $m$, where $z_{mk}\in\{1,...,J\}$ and $z_m$ represent the set of feature assignments in rule $m$. We sample $z_m$ from a  multinomial distribution with  weights $p$ drawn from a Dirichlet distribution parameterized by $\theta$.  Let $l_{mj}$ denote the number of items with attribute $j$ in rule $m$, i.e., $l_{mj} = \sum_k \mathbbm{1}(z_{mk} = j)$ and $\sum_{j}l_{mj} = L_m$. It means $l_{mj}$ items share the same feature $j$ and therefore can be merged into a condition. We truncate the multinomial distribution to only allow $l_{mj}\leq |\mathcal{V}_j|$.    %Let $V_{mj}$ denote the merged value set of this condition.

Here's the prior model, with shape parameters $H_s = \{\alpha_M, \beta_M, \alpha_L, \beta_L, \theta\}$.
\vspace{-2mm}
\begin{table}[h]\renewcommand{\arraystretch}{1.3}
\centering
\label{my-label}
\begin{tabular}{lll}
$M \sim \text{Poisson}(\lambda_M)$& $\lambda_M \sim \text{Gamma}(\alpha_M,\beta_M)$ \\

$L_m \sim \text{Poisson}(\lambda_L), \forall m $ & $ \lambda_L \sim \text{Gamma}(\alpha_L,\beta_L)$ \\
$z_{m}\sim \text{Multinomial}(p),\forall m  $ &$p\sim \text{Dirichlet}(\theta)$ 
%\\bottomrule
\end{tabular}
\end{table}
\vspace{-2mm}

Remarks: we use Multinomial-Dirichlet distribution for feature assignment for its clustering property of the outcomes. The prior model will tend to re-use features already in the rule. This is consistent with the interpretability goal of our model: to form a MARS model with fewer features so that multiple items can be merged in to one condition. The prior does not consider values in  each item since they do not affect the size and the shape of the model and therefore have no effect on the interpretability.
% In summary,
%the prior for MARS model follows a distribution below. C is a function of $H_s$ and $\Gamma(\cdot)$ is a gamma function. 
%\begin{align}
% p(R; H_s) \propto\frac{\Gamma(M^*+\alpha_M)C^M}{\Gamma(M^*+1)}\prod_{m=1}^M \frac{\Gamma(L_m+\alpha_L)}{\Gamma(L_m+1)} \cdot & \notag\\
%   \frac{\prod_{j=1}^J \Gamma(l_{mj}+\theta_j) }{(\beta_L+1)^{L_m}\Gamma(L_m + \sum_{j=1}\theta_j)} &\label{eqn:pR}     
%\end{align}

\paragraph{Conditional Likelihood}
Now we consider the predictive accuracy of a MARS by modeling the conditional likelihood of labels $\mathbf{y}$ given features $\mathbf{x}$ and a MARS model $R$. Our prediction on the outcomes are based on the coverage of MARS. According to formula (\ref{eqn:classifier}), if an observation satisfies $R$ (covered by $R$), it is predicted to be positive, otherwise, it's negative. We assume  label $y_n$ is drawn from Bernoulli distributions. Specifically, when $R(\mathbf{x}_n)=1$, i.e., $\mathbf{x}_n$ satisfies the rule set, $y_n$ has probability $\rho_+$ to be positive, and when $R(\mathbf{x}_n)=0$, $y_n$ has probability $\rho_-$ to be negative. 
$\rho_+, \rho_-$ govern the predictive accuracy on the training data. We assume that they are drawn from two Beta distributions  with  hyperparameters $(\alpha_+,\beta_+)$ and $(\alpha_-, \beta_-)$, respectively, which control the predictive power of the model.  The conditional likelihood is as below given parameters $H_c = \{\alpha_+,\beta_+,\alpha_-,\beta_-\}$:
\begin{equation}\label{eqn:pcond}
 p(\mathbf{y}|\mathbf{x}, R;H_c) \propto B(\text{tp}+\alpha_+,\text{fp}+\beta_+)B(\text{tn}+\alpha_-,\text{fn}+\beta_-),
\end{equation}
where tp, fp, tn and fn represent the number of true positives, false positives, true negatives and false negatives, respectively. $B(\cdot)$ is a Beta function and comes from integrating out $\rho_+, \rho_-$ in the conditional likelihood function. 
%\begin{align}
% \log  p(\mathbf{y}|&\mathbf{x}, R;H_c)=n_+ \log(\rho_+) + n_-\log (\rho_-)  \notag \\
% &-\sum_{i, R(\mathbf{x}_i)=1}^n \mathbbm{1}(R(\mathbf{x}_i) \neq y_i)\left(\frac{1-\rho_-}{\rho_+} \right) \notag \\
% &-\sum_{i,R(\mathbf{x}_i)=0}^n \mathbbm{1}(R(\mathbf{x}_i) \neq y_i)\left(\frac{1-\rho_+}{\rho_-} \right) ,\label{eqn:pcond}
%\end{align}
%where $\sum_{i, R(\mathbf{x}_i)=1}^n \mathbbm{1}(R(\mathbf{x}_i) \neq y_i)$ represents the false positive errors and  $\sum_{i}^n \mathbbm{1}(R(\mathbf{x}_i) \neq y_i)$ represents the false negative errors. $\rho_+, \rho_-$ control the costs for different type of errors. When $\rho_+ = \rho_- = \rho$, the conditional likelihood reduces to 
%\begin{equation}
%\log p(\mathbf{y}|\mathbf{x},R:H_c) = n\log \rho - \sum_{i}^n \mathbbm{1}(\hat{y}_i \neq y_i) \log \frac{1-\rho}{\rho}
%\end{equation}
%which is equivalent to minimizing the classification error.
%$B(\cdot)$ is a Beta function and comes from integrating out $\rho_+, \rho_-$ in the conditional likelihood function. 
 
We will write $p(R; H_s)$ as $p(R)$ and $ p(\mathbf{y}|\mathbf{x}, R;H_c)$ as $p(\mathbf{y}|\mathbf{x})$, ignoring dependence on parameters when necessary. Regarding setting hyperparameters $H_s, H_c$, there are natural settings for $\theta$ (all entries being 1). This means there's no prior preference for features. For Gamma distributions,  we set $\alpha_M$ and $\alpha_L$ to 1. Then the strength of the prior for constructing a simple MARS depends on $\beta_M$ and $\beta_L$. Increasing $\beta_M$ and $\beta_L$ decreases the expected number of rules and the expected length of rules, penalizing more on larger models. There are four real-valued parameters in the conditional likelihood to set, $\alpha_+, \beta_+,\alpha_-, \beta_-$. The ratio of $\alpha_+,\beta_+$ and $\alpha_-,\beta_-$ are associated with the expected predictive accuracy. Therefore we should always set $\alpha_+ >\beta_+, \alpha_- > \beta_-$. By default, $\beta_+, \beta_-$ are set to 1. Setting values of the parameters can be done through cross-validation, another layer of hierarchy with more diffuse hyperparameters, or plain intuition.

%\paragraph{Posterior}
%\begin{equation}
%p(R|S)\propto - \sum_{i}^n \mathbbm{1}(\hat{y}_i \neq y_i) \log \frac{1-\rho}{\rho} + p(R)
%\end{equation}

\subsection{Clustering of Features} We use Multinomial-Dirichlet in the prior model to take advantage of the ``clustering'' effect in feature assignment. Our goal is to formulate a model which favors rules with fewer features. Here we prove this effect.
 Let $R$ denote a MARS model and $l_{mj}$ represent the number of items in rule $m$ taking feature $j$. Now we do a small change in $R$: pick two features $j_1, j_2$ in rule $m$ where $l_{mj_1}\geq l_{mj_2}$ and  replace an item taking feature $j_2$ with an item taking feature $j_1$, and we denote the new rule set as $R^\prime$.  Every rule in $R^\prime$ remains the same as $R$ except in the $m$-th rule. Let $l_{mj_1}^\prime, l_{mj_2}^\prime$ denote the number of items taking feature $j_1$ and $j_2$ in the new model, and $l_{mj_1}^\prime = l_{mj_1}+1$ and $l_{mj_2} = l_{mj_2}^\prime - 1$. We claim this flip of feature increases the prior probability of the model, i.e., 
\begin{theorem}\label{thm:chinese_restaurant}
If $l_{mj_1}+\theta_{j_1}\geq l_{mj_2}+\theta_{j_2}$, then $p(R^\prime)\geq p(R)$.
\end{theorem}
This theorem states that MARS model favors rules using fewer features. For example, for  two rules   $[\text{state }=\text{California or Texas}]$ \emph{AND} $[\text{marital status}=\text{married}]$ and  $[\text{state}=\text{California}]$ \emph{AND} $[\text{marital status}=\text{married}]$ \emph{AND} $[\text{age}\geq 45]$, MARS will favor the former, if everything else in a rule set being equal. 
When we choose uniform prior where all $\theta_j$ are equal, the theorem will be reduced to a simpler form, that the model always tends to reuse the most prevalent features. 
(All proofs are in the supplementary material.)

\section{Inference Method}
%Our goal is to find a maximum \emph{a priori} (MAP) MARS model with  the posterior:
%\begin{equation}\label{eqn:posterior}
%    p(R|S;H_s,H_c)\propto p(R;H_s)p(\mathbf{y}|\mathbf{x}, R;H_c).
%\end{equation}
%We write $p(R|S;H_s,H_c)$ as $p(R|S)$, $p(R;H_s)$ as $p(R)$ and $p(\mathbf{y}|\mathbf{x}, R;H_c)$ as $p(\mathbf{y}|\mathbf{x})$ for simple notations, ignoring dependence on parameters $H_s, H_c$ when appropriate.
%
Inference for rule-based models is challenging because it involves a search over exponentially
many possible sets of rules: since each rule is a conjunction of conditions, the number of rules
increases exponentially with the number of features in a data set, and the solution space (all possible rule sets) is a powerset of the rule space. To obtain a maximum \emph{a posteriori} (MAP) model within this solution space, Gibbs Sampling takes tens of thousands of iterations or more to converge even searching within a reduced space of only a couple of thousands of pre-mined and pre-selected rules \cite{LethamRuMcMa15, WangRu15}. %We propose a more efficient inference algorithm here.

Here we propose a more efficient inference algorithm that adopts the basic search procedure in simulated annealing. 
Given an objective function $p(R|S)$ over discrete search space of different rule sets and a temperature 
schedule function over time steps, $T^{[t]} = T_0^{1 - \frac{t}{N_\text{iter}}}$, a simulated
annealing \cite{kirkpatrick1983optimization} procedure is a discrete time, discrete state Markov
Chain where at step $t$, given the current state $R^{[t]}$, the
next state $R^{[t+1]}$ is chosen by first proposing a neighbor and accepting it with  probability $\exp\left(\frac{p(R^{[t+1]}|S) - p(R^{[t]}|R)}{T^{[t]}}\right)$. 
In this framework, we incorporate following strategies for faster computation. 1) instead of randomly proposing a neighboring solution, we aim to improve from the current solution by evaluating neighbors and pick the right one to propose. 2) we use theoretical bounds for bounding the sampling chain to reduce computation. Below, we first derive bounds on MAP models and then detail the proposing step.% in the search procedure.

\subsection{Bounds on MAP models}
We exploit the model formulation to guide us in the search.
 We start by looking at MARS models with one rule removed. 
Removing a rule will yield a simpler model  but may lower the likelihood. However, we can prove that the loss in likelihood is bounded as a function of the support. For a rule set $R$, we use $R_{\backslash z}$ to represent a set that contains all rules from $R$ except the $z^\text{th}$ rule, i.e., $ R_{\backslash z} = \{r_m|r_m\in R, m\neq z\}.$ Let $N_+, N_-$ represent the number of positive and negative examples, respectively.
%Define $\Upsilon = \frac{1-\rho_-}{\rho_+}$ and Notate the support of a rule as $\text{supp}(r) = \sum_{n}r(\mathbf{x}_n).$
%Then the following holds: 
%\begin{lm}\label{lm:upsilon}
%$p(\mathbf{y}|\mathbf{x},R)\geq \Upsilon^{\text{\rm supp}(z)}p(\mathbf{y}|\mathbf{x},R_{\backslash z})$.
%\end{lm}
%$\Upsilon$ is meaningful if $\Upsilon \leq 1$, otherwise this lemma means adding a rule always increases the conditional likelihood. $\Upsilon \leq 1$ means false negative rate is less than true positive rate.
Define $$ \Upsilon = \frac{\beta_-(N_++\alpha_++\beta_+-1)}{(N_-+\alpha_-+\beta_-)(N_++\alpha_+-1)}, $$  Notate the support of a rule as $\text{supp}(r) = \sum_{n}r(\mathbf{x}_n).$
Then the following holds: 
\begin{lm}\label{lm:upsilon}
$p(\mathbf{y}|\mathbf{x},R)\geq \Upsilon^{\text{\rm supp}(z)}p(\mathbf{y}|\mathbf{x},R_{\backslash z})$.
\end{lm}
$\Upsilon$ is meaningful if $\Upsilon \leq 1$, otherwise this lemma means adding a rule always increases the conditional likelihood. This condition almost always holds since $\alpha_+>\beta_+, \alpha_->\beta_-$ and we do not set $\beta_+$ to a significantly large value. In practice it is recommended to set $\beta_+, \beta_-$ to 1. 

We introduce notations for later use.  Let $\mathcal{L}^*$ denote the maximum likelihood of dataset $S$, which is achieved when all data are classified correctly , i.e. TP $= N_+$, FP $= 0$, TN $= N_-$, and FN $= 0$, giving:
$\mathcal{L}^*: = B(N_++\alpha_+,\beta_+)B(N_-+\alpha_-,\beta_-)$.  Let
$v^{[t]}$ denote the best solution found until iteration $t$, i.e.,
\begin{equation*}
v^{[t]} = \max_{\tau \leq t}p(R^{[\tau]}|S).
\end{equation*}
According to the prior model, having too many rules penalizes the model due to the large complexity. Therefore, to hold a spot in the model, each rule needs to make enough ``contribution'' to the objective, i.e., capturing enough of the positive class, to cancel off the decrease in the prior.
Therefore, we claim that the support of rule in the MAP model is lower bounded, and the bound becomes tighter as $v^{[t]}$ increases along the iterations.
\begin{theorem}\label{thm:support}
Take a dataset $S$ and a MARS model with parameters $$H = \left\{\alpha_M, \beta_M, \alpha_L, \beta_L, \theta, \alpha_+, \beta_+, \alpha_-, \beta_-\right\}.$$  Define $R^* \in \arg\max_R p(R|S;H)$. If $\alpha_M<\beta_M, \alpha_L<\beta_L, \alpha_+>\beta_+, \alpha_->\beta_-$, we have: 
$$\forall r\in R^*, \text{supp}(r)\geq \left\lceil\frac{\log \frac{M^{[t]}+\alpha_M-1}{M^{[t]}\alpha_M\Omega}}{\log \Upsilon}\right\rceil,$$ and \vspace{-2mm}
$$|R^*| \leq M^{[t]} = \left\lfloor \frac{\log \mathcal{L}^* +\log p(\emptyset) -v^{[t]}}{\log \Omega } \right\rfloor,$$
where $\Omega = \frac{(\beta_M+1)(\beta_L+1)^{\alpha_L+1}\sum_{j=1}^J \theta_j}{ \alpha_M\beta_L^{\alpha_L}\alpha_L\max(\theta)} $.
\end{theorem}
$p(\emptyset)$ is the prior of an empty set. $\mathcal{L}^*$ and $p(\emptyset)$ upper bound the conditional likelihood and prior, respectively. The difference between $\log\mathcal{L}^* + \log p(\emptyset)$ and $v^{[t]}$, the numerator, represents the room for improvement from the current solution $v^{[t]}$. The smaller the difference, the smaller the $M^{[t]}$. When we choose $\alpha_M = 1$, the bound on support is reduced to  $$ \text{supp}(r)\geq\left\lceil\frac{\log\frac{1}{\Omega}}{\log \Upsilon}\right\rceil.$$
We can control the bounds by change parameters to increase or decrease $\Omega$.  As $\Omega$ increases, the bound $M^{[t]}$ decreases, which indicates a stronger preference for a simpler model with a smaller number of rules. Simultaneously, the lower bound for support increases, which is equivalent to reducing the search space. To increase $\Omega$, one can either decrease $\frac{\alpha_M}{\beta_M}$, the expected number of rules from the prior distribution, or decrease $\frac{\alpha_L}{\beta_L}$, the expected number of items in each rule.
 
% Then we study when it becomes too costive to add a value to a feature.
%\begin{theorem}\label{thm:value}
%    \begin{equation*}
%s \geq \frac{\log\frac{(L_m+1)(\beta_L+1)}{L_m+\alpha_L}\frac{L_m + \sum_j\theta_j}{l_{mj^\prime}+\theta_{j^\prime}}}{\log \Upsilon}
%\end{equation*}
%\end{theorem}
%
%This prevents overfitting !!

We incorporate the bound on support in the search to check if a rule qualifies to be included. 

\subsection{Proposing steps}
Here we detail the proposing step at each iteration in the search algorithm.
We simultaneously define the set of neighbors and a proposal for a neighbor.
A ``next state'' is proposed by first selecting an action to alter the current MARS and then choose from the ``neighboring'' models generated by that action. To improve search efficiency, we do not perform a random action, but instead, we sample from misclassified examples to determine an action that can improve the current state $R^{[t]}$. 
If the misclassified example is positive, it means the $R^{[t]}$ fails to ``cover'' it and therefore needs to increase the coverage by randomly choosing one of the following actions.
\begin{itemize}
\item \emph{Add a value}: Choose a rule $r_m \in R^{[t]}$, a condition $c_k \in r_m$ and then a candidate value $v\in \mathcal{V}_{z_{mk}} \backslash \nu(c_k)$, then $c_k \leftarrow (z_{mk}, \nu(c_k)\cup v)$. 
\item \emph{Remove a condition}: Choose a rule $r_m \in R^{[t]}$ and a condition $c_k \in r_m$, then $r_m = \{c_{k^\prime} \in r_m |c_{k^\prime}\neq c_k\}$
\item \emph{Add a rule}: Generate a new rule $r^\prime$ with $\text{supp}(r^\prime)$ bounded by Theorem~\ref{thm:support}, $R^{[t+1]} \leftarrow R^{[t]}\cup r^\prime$
\end{itemize}
where we use $\nu(\cdot)$ to access the feature in a condition.  The three actions above increase the coverage of a rule set.

\begin{figure*}[b!]
\centering
\subfloat[Average error on hold-out sets.]{
  \includegraphics[width=0.29\textwidth,trim={0 0 0 0.75cm},clip]{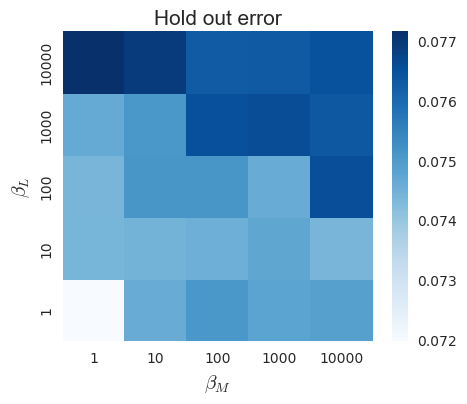}
}
\subfloat[Average number of conditions.]{
  \includegraphics[width=0.27\textwidth,trim={0 0 0 0.75cm},clip]{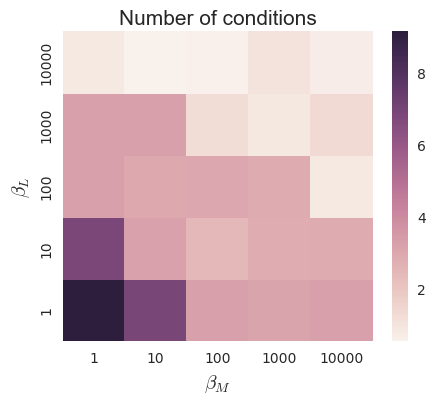}
}
\subfloat[Average number of features.]{
  \includegraphics[width=0.28\textwidth,trim={0 0 0 0.75cm},clip]{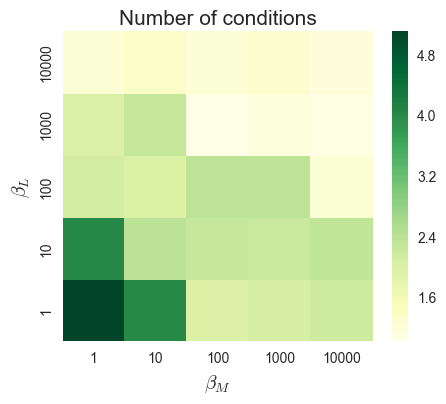}
}
\caption{Effect of shape parameters on predictive accuracy and interpretability}\label{fig:param}
\end{figure*}

On the other hand, if the misclassified example is negative, it means $R^{[t]}$ covers more than it should and therefore needs to reduce the coverage by randomly choosing one of the following actions.
\begin{itemize}
\item \emph{Add a condition}: Choose a rule $r_m\in R^{[t]}$ first, choose a feature $j^\prime \in \{1,\cdots,J\}\backslash z_m$ and then a set of values $V^\prime \in \mathcal{V}_{j^\prime}$, then update $r_m \leftarrow r_m \cup (j^\prime, V^\prime)$
\item \emph{Remove a rule}: Choose a rule $r_m \in R^{[t]}$, then $R^{[t+1]} = \{r \in R^{[t]}|r\neq r_m\}$ 
\end{itemize}
The two actions above reduce the coverage of a model.

Performing an action involves choosing a value, a condition, or a rule. Different choices result in different neighboring models. To select one from them, we evaluate the posterior on every model.  Then a choice is made between exploration (choosing a random model) and exploitation (choosing the best model). This randomness helps to avoid local minima and helps the Markov Chain to converge to a global optimum.  

See the algorithm in Algorithm 1.
\begin{algorithm}[h!]
\caption{\label{myalg} Inference algorithm. %Theorem \ref{thm:supportb} (or 2) determines $C^{[t]}$ %\textcolor{red}{Which model is this for if it only uses Theorem 1? What about Theorem 2?}
}
\begin{algorithmic}\label{alg:smart}
%\Procedure{Simulated Annealing}{$N_\text{iter}$}
\State $R^{[0]} \leftarrow$ a randomly generated rule set
%\State $A_\text{max} \leftarrow R^{[0]}$ 
 \For{$t=0\rightarrow N_\text{iter}$}
     \State $(\mathbf{x}_k,y_k) \leftarrow$ a random example drawn from data points misclassified by  $R^{[t]}$ 
     \If{$y_k =1$}
%     \State $q = \frac{1}{2}e^{-\max\{0,|R^{[t]}| - \sum_{l=1}^L m_l^{[t]}\}}$
\State $R^{[t+1]} \leftarrow \begin{cases}
        \text{AddValue}(R^{[t]}),\text{ w/ prob } \frac{1}{3} \\
        \text{RemoveCondition}(R^{[t]}),\text{ w/ prob}\frac{1}{3} \\
        \text{AddRule}(R^{[t]}),\text{ w/ prob }\frac{1}{3}\text{ (using}\\
        \quad\quad\quad\quad\quad\quad\quad\quad\text{Theorem 2)}
        \end{cases}$
        \Else
\State $R^{[t+1]}  \leftarrow \begin{cases}
        \text{AddCondition}(R^{[t]}),\text{ w/ prob }\frac{1}{2} \\
        \text{RemoveRule}(R^{[t]}),\text{ w/ prob }\frac{1}{2} \\
        \end{cases}$
        \EndIf
\State $R_\text{max} \leftarrow \arg\max(p(R_\text{max}|S),p(R^{[t+1]}|S))$ (Check for improved optimal solution)
\State $\alpha = \min\left\{1,\exp\left(\frac{p(R^{[t+1]}|S) - p(R^{[t]}|S)}{T^{[t]}}\right)\right\}$ (Probability of an annealing move)
\State  $R^{[t+1]}  \leftarrow R^{[t+1]}, \text{  with probability }\alpha $
\EndFor
\State return $R_{\text{max}}$
%\EndProcedure
\end{algorithmic}
\end{algorithm}

\section{Experimental Evaluation}
We perform a detailed experimental evaluation of MARS models on simulated and real-world data sets. The first part of our experiments is designed to study the effect of hyperparameters on the interpretability and predictive accuracy. The second part of the experiments compare MARS with classic and state-of-the-art benchmark baselines. In the last part of the experiments, we conduct a usability study with humans to study how quickly and correctly humans understand and use different rule-based models.

\subsection{Accuracy \& Interpretability Trade-off}
%We study the effect of the shape parameters on the interpretability and predictive accuracy of the model. 
We generate 10 data sets of 100k observations with 50 arbitrary numerical features uniformly drawn from 0 to 1. For each dataset, we construct a set of 10  rules by first drawing the number of conditions uniformly from 1 to 10 for each rule and then filling conditions with randomly selected features. Since the data are numerical, we generate a range for each feature by randomly selecting two values from 0 to 1, one as lower boundary and the other as upper boundary. These ten rules are the ground truth rule set denoted as $R^*$. Then we generate labels $\mathbf{y}$ from $R^*$: observations that satisfy $R^*$ are positive. Then each dataset is partitioned into 75\% training and 25\% testing. To apply MARS 
model, we discretize each feature into ten intervals and obtain a binary dataset of size 100k by 500 on which we run the proposed model.  We set entries in $\theta$ to 1, $\alpha_+ = \alpha_- = 100$ and $\beta_+ = \beta_- = 1$. Out of the four shape parameters $\alpha_M, \beta_M, \alpha_L, \beta_L$, we fix $\alpha_M, \alpha_L$ to 1 and only vary $\beta_M,\beta_L$. Larger values of $\beta_M,\beta_L$ indicate a stronger prior preference for simpler models. Let $\beta_M, \beta_L$ take values from $\{1,10,100,1000,10000\}$, giving a total of 25 sets of parameters. On each training data set we run MARS model with 25 sets of parameters and then evaluate the output model on the test set. We report in Figure~\ref{fig:param} the hold-out error, the number of conditions and the number of features used in the model. Each block corresponds to a  parameter set. The values are averaged over ten datasets.

In all three figures above, smaller values are represented with lighter colors, indicating either lower errors or smaller complexities. 
The left-bottom corner represents models with least constraint on complexity ($\beta_M=1, \beta_L=1)$ and they achieve the lowest error but at the cost of highest complexity and largest feature set.
As $\beta_M$ and $\beta_L$ increase, the model becomes less complex, with fewer conditions and fewer feature, but at the cost of predictive accuracy. The right-top corner represents models with the strongest preference for simplicity: smallest model but largest error.
The three figures show a clear pattern of the trade-off between interpretability and predictive accuracy. %When applying to real datasets, it is recommended to set aside a validation set for tuning the parameters.

\subsection{Real World Datasets}
We then evaluate the performance of MARS on four real-world datasets and compare the performance with classic and state-of-the-art rule-based models. We also apply black-box models random forest and XGBoost to these datasets as a benchmark to quantify the possible loss (if any) in predictive accuracy for gaining interpretability.

\textbf{Datasets} We analyze four real-world datasets. 1) \emph{Juvenile}\cite{osofsky1995effect} (4023 observations and 69 reduced features), to study the consequences of juvenile exposure to violence. %The dataset was collected via a survey sent to juveniles.   
2) \emph{credit card} (30,000 observations and 24 features), to predict the default of credit card payment \cite{yeh2009comparisons}. 3) \emph{census} \cite{kohavi1996scaling}(48,842 observations and 14 features)  to predict the annual income based on individual's demographic information. 4) \emph{recidivism} (11,645 observations and 106 features). Data were collected on offenders who were
sentenced in 1986 and committed one or more felony crimes. All datasets contain both categorical and numerical attributes. We discretized numerical features to 10 intervals. 

\begin{table*}[h!]\renewcommand{\arraystretch}{1.1}
\footnotesize
\centering
\caption{Evaluation of predictive performance and model complexity over 5-fold cross validation}\label{tab:exp}
\label{my-label}
\begin{tabular}{@{\hskip3pt}l@{\hskip3pt}@{\hskip3pt}c@{\hskip3pt}@{\hskip3pt}c@{\hskip3pt}@{\hskip3pt}c c@{\hskip3pt}@{\hskip3pt}c@{\hskip3pt}@{\hskip3pt}cc@{\hskip3pt}@{\hskip3pt}c@{\hskip3pt}@{\hskip3pt}cc@{\hskip3pt}@{\hskip3pt}c@{\hskip3pt}@{\hskip3pt}c@{\hskip3pt}}
\toprule
Task   & \multicolumn{3}{@{\hskip3pt}c@{\hskip3pt}}{Juvenile }& \multicolumn{3}{@{\hskip3pt}c@{\hskip3pt}}{Credit card } & \multicolumn{3}{@{\hskip3pt}c@{\hskip3pt}}{ Census} & \multicolumn{3}{@{\hskip3pt}c@{\hskip3pt}}{Recidivism}  \\ \hline
Method & accuracy     &   $n_\text{cond}$            &   $n_\text{feat}$            & accuracy             &   $n_\text{cond}$       &   $n_\text{feat}$         &  accuracy            &     $n_\text{cond}$         &  $n_\text{feat}$    &accuracy            &     $n_\text{cond}$         &  $n_\text{feat}$     \\ 
Ripper  &  .88(.01)             &  35(13)            & 23(5)             &   .82(.01)         &   23(8)&    12(2)      &   .84(.01)    &  67(11)           &  7(0)          &.78(.00)&78(18)&32(4)     \\ 
CBA    & .88(.01)              &  27(22)             &     18(12)        &   .80(.01)         &    35(3)         &    6(0)      &   .79(.01)          &       13(12)        &   6(2)         &.72(.01)&87(25)& 27(5)    \\ 
SBRL    & .88(.01)              & 10(2)             & 9(2)             &   .82(.00)          &   15(2)         & 10(2)       &  .82(.00)         &     32(2)       &    10(1)        &.75(.00)&10(1)& 9(1)    \\ 
BRS    &  .88(.01)            & 21(4)             &  11(3)             &      .81(.01)       & 17(2)           &  8(2)          &   .79(.01)           &33(11)            & 11(2)      &.79(.01)&33(11)&  19 (3)     \\ 
\textbf{MARS}    &   .89(.00)           &    18(3)           &   6(2)           & .82(.01) &   10(7)          &  5(3)       &  .80(.00)          &   14(8)           &     5(3)        &.74(.02)& 6(3)&  3(1) \\
RF    &   .90(.00)          &  --            &   --          & .82(.00)    &     --     &    --     & .86(.00)         &       --     &       --      &.96(.00)&&    \\ 
XGBoost &  .91(.01)         &  --            &   --          &  .83(.01)   &     --     &    --     & .87(.01)       &       --     &       --       &.79(.05)&--& --  \\ \bottomrule
\end{tabular}
\end{table*}
\normalsize

\textbf{Baselines} We benchmark the performance of MARS against the following rule-based models for classification: Scalable Bayesian Rule Lists (SBRL) \cite{yang2016scalable}, Classification Based on Associations (CBA) \cite{ma1998integrating}, Repeated Incremental Pruning to Produce Error Reduction (Ripper) \cite{cohen1995fast} and Bayesian Rule Sets (BRS) \cite{wangbayesian}. CBA and Ripper were designed
to bridge the gap between association rule mining and classification 
and thus focused mostly on optimizing for predictive accuracy. They are among the earliest and most-cited work on rule-based classifiers.
On the other hand, BRS and SBRL, two recently proposed frameworks
aim to achieve simpler models alongside predictive accuracy. In addition to rule-based models, we also compare with random forests and XGBoost to benchmark the performance without accounting for interpretability. 

\textbf{Experimental Setup} We performed 5-fold cross validation for each method. The MARS model  has a set of hyperparameters $\alpha_M, \beta_M, \alpha_L, \beta_L, \theta, \alpha_+, \beta_+$, $\alpha_-$ and $\beta_-$.  We set entries in $\theta$ to 1, $\alpha_+ = \alpha_- = 100$ and $\beta_+ = \beta_- = 1$. $\alpha_M,\beta_M$ control the number of rules and $\alpha_L,\beta_L$ control lengths of rules.  We set $\alpha_M, \alpha_L$ to 1 and vary $\beta_M, \beta_L$. We set aside 20\% of data during training for parameter tuning and used a grid search to locate the best set of parameters. We use R and python packages for random forest, SBRL, CBA and Ripper \cite{Rweka} and use the publicly available code for BRS. Due to the computation limit, SBRL cannot handle rules longer than 3 for these datasets, so we set the maximum rule length to 2. The parameters are tuned within each fold to obtain the optimal model for each method.

%
%\begin{table}[h]\renewcommand{\arraystretch}{1.1}
%\centering
%\caption{Evaluation of predictive performance and model complexity}\label{tab:exp}
%\label{my-label}
%\begin{tabular}{l|c|c|c|c|c|c|c|c|c}
%\toprule
%Task   & \multicolumn{3}{c|}{test accuracy} & \multicolumn{3}{c|}{\# of conditions } & \multicolumn{3}{c}{ \# of unique features} \\ \toprule
%Method & Juvenile     &   Credit            &   Adult            & Juvenile             &    Credit       &   Adult         &  Juvenile            &    Credit         &   Adult         \\ 
%Ripper  &  0.88             &  0.81             & 0.84              &     34.8          &    22.6&    67.2        &   8.0    &  8.0            &  7.0            \\ 
%SBRL    & 0.88              & 0.81              &  0.82             &   26.0          &   14.8         &  32.0          &  8.6            &     9.8         &    10.2          \\ 
%BRS    &  0.88            & 0.81              &  0.74             &             & 17.2           &  10.4          &              & 8.2             & 6.2             \\ 
%CBA    & 0.87              &  0.79             &     0.79          &            &    20.5         &     14.5       &             &       13.5        &   6.8           \\ 
%\textbf{MARS}    &   0.90           &    0.82            &  0.81             &             11.8 &   11.2          &  14.4        &      4.4       &   4.4           &   5.0           \\ \bottomrule
%\end{tabular}
%\end{table}

\textbf{Evaluations}
We evaluate the predictive performance and interpretability performance by measuring three metrics: i) the error rate on the test set, ii) the total number of conditions in the output model, and iii) the average number of unique features in the model. The three metrics are computed from the 5 folds, and we report the mean and standard deviation in Table~\ref{tab:exp}.
 
MARS achieved consistently competitive predictive accuracy using significantly fewer conditions and fewer features. On dataset \emph{juvenile} and \emph{creditcard}, MARS is the best performing model, highest accuracy, smallest complexity, and fewest features. On \emph{census} data, Ripper achieved higher accuracy 67 conditions, almost five times as many conditions used by MARS, and SBRL uses 32 conditions, while MARS uses only 14 on average. On all four datasets, MARS model uses the smallest number of features, even for juvenile dataset where MARS has more conditions than SBRL model but still wins over in the number of features. Interpretable models do lose predictive accuracy compared to black-box models, to trade for interpretability.

We are interested to see a MARS model and inspect if the grouping of categories is meaningful. We show a MARS model learned from dataset \emph{juvenile}. For demonstration purpose, we tune shape parameters to obtain a smaller model shown below. This MARS model achieves an accuracy of 0.86. It consists of two rules, and if a teenager satisfies either of them, then the model predicts the teenager will conduct delinquency in the future. In this dataset, features are questions and feature values are answer options.

1: [Have your friends ever hit or threatened to hit someone without any reason? = ``All of them'' or ``Not sure'' or ``Refused to Answer'']

2: [Have your friends purposely damaged or destroyed property that did not belong to them? = ``All of them'' \textbf{or} ``Most of them'' \textbf{or}  ``Some of them''] \emph{AND} [Did any of your family members use hard drugs? = ``Yes''] \emph{AND} [``Has any of your family members of friends ever beat you up with their fists so hard that you were hurt pretty bad? = ``Yes'']

It is interesting to notice that MARS grouped three values in the first rule together and considered they are interchangeable. This is intuitive to explain with common sense. People avoid answering when they feel alerted or uncomfortable with the question \cite{chisnall1993questionnaire,willis2004cognitive}. In this case, this question concerns the privacy of their friends, making people more reserved providing a definite answer. So they'd rather say not sure or refuse to answer than say yes.

\subsection{Interpretability Evaluation by Humans}
To further evaluate the model interpretability, in addition to quantitively measuring the size of the model, we would like to understand how quickly and how correctly humans understand a machine learning model. We designed a short survey and sent it to a group of 70 undergraduate students. The survey was designed as an online quiz with credit to motivate students to do it as accurately as possible. The students have been enrolled in a machine learning class for a couple of weeks and have some knowledge about predictive models.

We chose to show models built from dataset ``credit card'' since output models are smallest compared to other datasets so it's easier for humans to understand.  The students were asked to use the models to do predictions on given instances. Every method has 5 models, each from one of the 5 folds and a student is randomly shown one of them. Therefore, each student is shown one model for each five methods. The survey first taught them how to use a model with instructions and an example, and then asks them to use the model to make predictions on two instances\footnote{See the supplementary material for an example of teaching and using a model in the survey}. Their answers and response time are recorded.

Since all competing methods are rule-based models, it is important that students understand the notion of rules before working with any of the models. Therefore, we designed a screening question on rules and students can only proceed with the survey if they answer the question correctly. 66 students passed the test. 

\begin{figure}[h!]
\centering
  \includegraphics[width=0.47\textwidth]{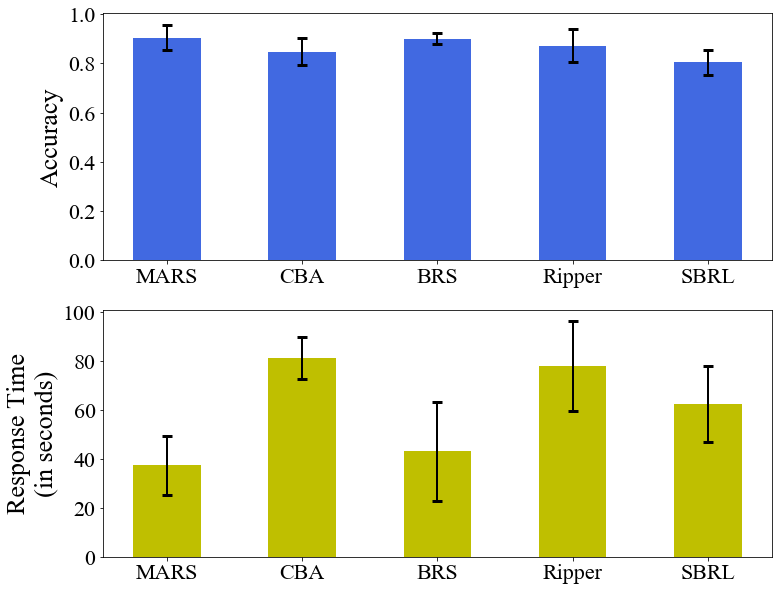}
\caption{Effect of shape parameters on predictive accuracy and interpretability}\label{fig:study}
\end{figure}

We report the accuracy and response time of each method averaged over 5 folds with Figure~\ref{fig:study}. (Note that response time refers to the total time for understanding the model and using the model to predict two instances.) Methods MARS and BRS achieve the highest accuracy, and SBRL achieves the lowest accuracy. We hypothesis this is because  SBRL uses an ordered set of rule connected by ``else-if'' which makes it a little more difficult to understand compared to un-ordered rules in the other four methods. For the response time, MARS uses a significantly small amount of time, less than half of that of CBA and Ripper, due to the Bayesian prior to favor small models and a concise presentation allowing multiple conditions in a rule. BRS also takes a very short time, a bit longer than MARS, followed by SBRL. MARS, BRS, and SBRL all have a Bayesian component to favor small models while CBA and Ripper do not, thus taking significantly longer to understand and use.

\section{Conclusions}
We proposed a Multi-value Rule Set (MARS) model which provides a more concise and feature-efficient model form to classify and explain. We developed an inference algorithm that incorporates theoretically grounded bounds to reduce computation. Compared with classic and state-of-the-art rule-based models, MARS showed competitive predictive accuracy while achieving a significant reduction in complexity and feature sets, thus improving the interpretability.
Our analysis on parameter tuning demonstrated a clear pattern in the trade-off between predictive accuracy and interpretability.
One major contribution of this work is that we offered a fresh perspective on the form of rules. Specifically, allowing multiple values in a condition provides a more concise presentation. The usability study with humans show that this form of model is easy to understand and use. Another contribution is that we demonstrated the possibility of using fewer features without hurting too much (if any) predictive performance, which is also an important aspect in interpretability alongside complexity. We believe the potential in the proposed multi-value rules is not just limited to MARS. They can be easily adopted in other rule-based models without changing other components in the model.

%
%\Appendix
\begin{spacing}{0.96}{
\bibliographystyle{abbrv-fr}      % basic style, author-year citations
\bibliography{nips_msr} }

\begin{thebibliography}{10}
\expandafter\ifx\csname fonteauteurs\endcsname\relax
\def\fonteauteurs{\scshape}\fi

\bibitem{chi1996fuzzy}
Z.~\bgroup\fonteauteurs\bgroup Chi\egroup\egroup{},
  H.~\bgroup\fonteauteurs\bgroup Yan\egroup\egroup{} et
  T.~\bgroup\fonteauteurs\bgroup Pham\egroup\egroup{} :
\newblock {\em Fuzzy algorithms: with applications to image processing and
  pattern recognition}, vol.~10.
\newblock World Scientific, 1996.

\bibitem{chisnall1993questionnaire}
P.~M. \bgroup\fonteauteurs\bgroup Chisnall\egroup\egroup{} :
\newblock Questionnaire design, interviewing and attitude measurement.
\newblock {\em Journal of the Market Research Society},
  35(4)\string:\penalty500\relax 392--393, 1993.

\bibitem{cohen1995fast}
W.~W. \bgroup\fonteauteurs\bgroup Cohen\egroup\egroup{} :
\newblock Fast effective rule induction.
\newblock \emph{In} {\em Proceedings of the twelfth international conference on
  machine learning}, p. 115--123, 1995.

\bibitem{Rweka}
K.~\bgroup\fonteauteurs\bgroup Hornik\egroup\egroup{},
  C.~\bgroup\fonteauteurs\bgroup Buchta\egroup\egroup{} et
  A.~\bgroup\fonteauteurs\bgroup Zeileis\egroup\egroup{} :
\newblock Open-source machine learning: {R} meets {Weka}.
\newblock {\em Computational Statistics}, 24(2)\string:\penalty500\relax
  225--232, 2009.

\bibitem{ishibuchi2001effect}
H.~\bgroup\fonteauteurs\bgroup Ishibuchi\egroup\egroup{} et
  T.~\bgroup\fonteauteurs\bgroup Nakashima\egroup\egroup{} :
\newblock Effect of rule weights in fuzzy rule-based classification systems.
\newblock {\em IEEE Transactions on Fuzzy Systems},
  9(4)\string:\penalty500\relax 506--515, 2001.

\bibitem{kirkpatrick1983optimization}
S.~\bgroup\fonteauteurs\bgroup Kirkpatrick\egroup\egroup{}, C.~D.
  \bgroup\fonteauteurs\bgroup Gelatt\egroup\egroup{}, M.~P.
  \bgroup\fonteauteurs\bgroup Vecchi\egroup\egroup{} \emph{et~al.} :
\newblock Optimization by simulated annealing.
\newblock {\em science}, 220(4598)\string:\penalty500\relax 671--680, 1983.

\bibitem{kohavi1996scaling}
R.~\bgroup\fonteauteurs\bgroup Kohavi\egroup\egroup{} :
\newblock Scaling up the accuracy of naive-bayes classifiers: A decision-tree
  hybrid.
\newblock \emph{In} {\em KDD}, vol.~96, p. 202--207. Citeseer, 1996.

\bibitem{lakkaraju2016interpretable}
H.~\bgroup\fonteauteurs\bgroup Lakkaraju\egroup\egroup{}, S.~H.
  \bgroup\fonteauteurs\bgroup Bach\egroup\egroup{} et
  J.~\bgroup\fonteauteurs\bgroup Leskovec\egroup\egroup{} :
\newblock Interpretable decision sets: A joint framework for description and
  prediction.
\newblock \emph{In} {\em ACM SIGKDD}, p. 1675--1684. ACM, 2016.

\bibitem{LethamRuMcMa15}
B.~\bgroup\fonteauteurs\bgroup Letham\egroup\egroup{},
  C.~\bgroup\fonteauteurs\bgroup Rudin\egroup\egroup{}, T.~H.
  \bgroup\fonteauteurs\bgroup McCormick\egroup\egroup{},
  D.~\bgroup\fonteauteurs\bgroup Madigan\egroup\egroup{} \emph{et~al.} :
\newblock Interpretable classifiers using rules and bayesian analysis: Building
  a better stroke prediction model.
\newblock {\em The Ann of Appl Stats}, 9(3)\string:\penalty500\relax
  1350--1371, 2015.

\bibitem{li2001cmar}
W.~\bgroup\fonteauteurs\bgroup Li\egroup\egroup{},
  J.~\bgroup\fonteauteurs\bgroup Han\egroup\egroup{} et
  J.~\bgroup\fonteauteurs\bgroup Pei\egroup\egroup{} :
\newblock Cmar: Accurate and efficient classification based on multiple
  class-association rules.
\newblock \emph{In} {\em ICDM}, p. 369--376. IEEE, 2001.

\bibitem{ma1998integrating}
B.~L. W. H.~Y. \bgroup\fonteauteurs\bgroup Ma\egroup\egroup{} et
  B.~\bgroup\fonteauteurs\bgroup Liu\egroup\egroup{} :
\newblock Integrating classification and association rule mining.
\newblock \emph{In} {\em KDD}, 1998.

\bibitem{malioutov2013exact}
D.~\bgroup\fonteauteurs\bgroup Malioutov\egroup\egroup{} et
  K.~\bgroup\fonteauteurs\bgroup Varshney\egroup\egroup{} :
\newblock Exact rule learning via boolean compressed sensing.
\newblock \emph{In} {\em International Conference on Machine Learning}, p.
  765--773, 2013.

\bibitem{micci2001preprocessing}
D.~\bgroup\fonteauteurs\bgroup Micci-Barreca\egroup\egroup{} :
\newblock A preprocessing scheme for high-cardinality categorical attributes in
  classification and prediction problems.
\newblock {\em ACM SIGKDD Explorations Newsletter},
  3(1)\string:\penalty500\relax 27--32, 2001.

\bibitem{miller1956magical}
G.~A. \bgroup\fonteauteurs\bgroup Miller\egroup\egroup{} :
\newblock The magical number seven, plus or minus two: some limits on our
  capacity for processing information.
\newblock {\em Psychological review}, 63(2)\string:\penalty500\relax 81, 1956.

\bibitem{osofsky1995effect}
J.~D. \bgroup\fonteauteurs\bgroup Osofsky\egroup\egroup{} :
\newblock The effect of exposure to violence on young children.
\newblock {\em American Psychologist}, 50(9)\string:\penalty500\relax 782,
  1995.

\bibitem{papaxanthos2016finding}
L.~\bgroup\fonteauteurs\bgroup Papaxanthos\egroup\egroup{},
  F.~\bgroup\fonteauteurs\bgroup Llinares-Lopez\egroup\egroup{},
  D.~\bgroup\fonteauteurs\bgroup Bodenham\egroup\egroup{} et
  K.~\bgroup\fonteauteurs\bgroup Borgwardt\egroup\egroup{} :
\newblock Finding significant combinations of features in the presence of
  categorical covariates.
\newblock \emph{In} {\em NIPS}, p. 2271--2279, 2016.

\bibitem{rijnbeek2010finding}
P.~R. \bgroup\fonteauteurs\bgroup Rijnbeek\egroup\egroup{} et J.~A.
  \bgroup\fonteauteurs\bgroup Kors\egroup\egroup{} :
\newblock Finding a short and accurate decision rule in disjunctive normal form
  by exhaustive search.
\newblock {\em Machine learning}, 80(1)\string:\penalty500\relax 33--62, 2010.

\bibitem{ritchie2015methods}
M.~D. \bgroup\fonteauteurs\bgroup Ritchie\egroup\egroup{}, E.~R.
  \bgroup\fonteauteurs\bgroup Holzinger\egroup\egroup{},
  R.~\bgroup\fonteauteurs\bgroup Li\egroup\egroup{}, S.~A.
  \bgroup\fonteauteurs\bgroup Pendergrass\egroup\egroup{} et
  D.~\bgroup\fonteauteurs\bgroup Kim\egroup\egroup{} :
\newblock Methods of integrating data to uncover genotype-phenotype
  interactions.
\newblock {\em Nature Reviews Genetics}, 16(2)\string:\penalty500\relax 85--97,
  2015.

\bibitem{pmlr-v56-Tran16}
T.~\bgroup\fonteauteurs\bgroup Tran\egroup\egroup{},
  W.~\bgroup\fonteauteurs\bgroup Luo\egroup\egroup{},
  D.~\bgroup\fonteauteurs\bgroup Phung\egroup\egroup{},
  J.~\bgroup\fonteauteurs\bgroup Morris\egroup\egroup{},
  K.~\bgroup\fonteauteurs\bgroup Rickard\egroup\egroup{} et
  S.~\bgroup\fonteauteurs\bgroup Venkatesh\egroup\egroup{} :
\newblock Preterm birth prediction: Stable selection of interpretable rules
  from high dimensional data.
\newblock \emph{In} {\em Proceedings of the 1st Machine Learning for Healthcare
  Conference}, vol.~56 de {\em Proceedings of Machine Learning Research}, p.
  164--177, Northeastern University, Boston, MA, USA, 18--19 Aug 2016. PMLR.

\bibitem{WangRu15}
F.~\bgroup\fonteauteurs\bgroup Wang\egroup\egroup{} et
  C.~\bgroup\fonteauteurs\bgroup Rudin\egroup\egroup{} :
\newblock Falling rule lists.
\newblock \emph{In} {\em AISTATS}, 2015.

\bibitem{wang2017bayesian}
T.~\bgroup\fonteauteurs\bgroup Wang\egroup\egroup{},
  C.~\bgroup\fonteauteurs\bgroup Rudin\egroup\egroup{},
  F.~\bgroup\fonteauteurs\bgroup Doshi\egroup\egroup{},
  Y.~\bgroup\fonteauteurs\bgroup Liu\egroup\egroup{},
  E.~\bgroup\fonteauteurs\bgroup Klampfl\egroup\egroup{} et
  P.~\bgroup\fonteauteurs\bgroup MacNeille\egroup\egroup{} :
\newblock A bayesian framework for learning rule sets for interpretable
  classification.
\newblock {\em Journal of Machine Learning Research}, 2017.

\bibitem{wangbayesian}
T.~\bgroup\fonteauteurs\bgroup Wang\egroup\egroup{},
  C.~\bgroup\fonteauteurs\bgroup Rudin\egroup\egroup{},
  F.~\bgroup\fonteauteurs\bgroup Velez-Doshi\egroup\egroup{},
  Y.~\bgroup\fonteauteurs\bgroup Liu\egroup\egroup{},
  E.~\bgroup\fonteauteurs\bgroup Klampfl\egroup\egroup{} et
  P.~\bgroup\fonteauteurs\bgroup MacNeille\egroup\egroup{} :
\newblock Bayesian rule sets for interpretable classification.
\newblock {\em ICDM}, 2016.

\bibitem{willis2004cognitive}
G.~B. \bgroup\fonteauteurs\bgroup Willis\egroup\egroup{} :
\newblock {\em Cognitive interviewing: A tool for improving questionnaire
  design}.
\newblock Sage Publications, 2004.

\bibitem{ynormalize_addang2016scalable}
H.~\bgroup\fonteauteurs\bgroup Yang\egroup\egroup{},
  C.~\bgroup\fonteauteurs\bgroup Rudin\egroup\egroup{} et
  M.~\bgroup\fonteauteurs\bgroup Seltzer\egroup\egroup{} :
\newblock Scalable bayesian rule lists.
\newblock {\em ICML}, 2017.

\bibitem{yeh2009comparisons}
I.-C. \bgroup\fonteauteurs\bgroup Yeh\egroup\egroup{} et C.-h.
  \bgroup\fonteauteurs\bgroup Lien\egroup\egroup{} :
\newblock The comparisons of data mining techniques for the predictive accuracy
  of probability of default of credit card clients.
\newblock {\em Expert Systems with Applications},
  36(2)\string:\penalty500\relax 2473--2480, 2009.

\bibitem{yin2003cpar}
X.~\bgroup\fonteauteurs\bgroup Yin\egroup\egroup{} et
  J.~\bgroup\fonteauteurs\bgroup Han\egroup\egroup{} :
\newblock Cpar: Classification based on predictive association rules.
\newblock \emph{In} {\em SIAM International Conference on Data Mining}, p.
  331--335. SIAM, 2003.

\end{thebibliography}
\end{spacing}
\end{document}